# Ten ways to fool the masses with machine learning

**Fayyaz Minhas[1], Amina Asif[2], and Asa Ben-Hur[3]**

[1,2]PIEAS Data Science Lab, Pakistan Institute of Engineering and Applied Sciences, Islamabad, Pakistan

[3]Department of Computer Science, Colorado State University, Fort Collins, Colorado, USA

[1]afsar@pieas.edu.pk, [2]a.asif.shah01@gmail.com, [3]asa@cs.colostate.edu

**Abstract**

*"If you want to tell people the truth, make them laugh, otherwise they'll kill you". (source unclear)*

Machine learning and deep learning are the technologies of the day for developing intelligent automatic systems. However, a key hurdle for progress in the field is the literature itself: we often encounter papers that report results that are difficult to reconstruct or reproduce, results that mis-represent the performance of the system, or contain other biases that limit their validity. In this semi-humorous article, we discuss issues that arise in running and reporting results of machine learning experiments. The purpose of the article is to provide a list of "watch out!" points for researchers to be aware of when developing machine learning models or writing and reviewing machine learning papers.

**Introduction**

Machine learning, or learning using empirical data, is one of the most rapidly developing fields in computer science. It has been applied to a vast variety of problems and is currently ranked as one of the top fields in the Gartner hype cycle [1]. Typically, machine learning is aimed at solving a real-world problem (e.g., predicting whether a pizza joint is likely to have "good" pizza). To solve the problem, the machine learning engineer collects some "examples" (pizzas in our case) and gets them annotated (e.g., by food critics). Features are than extracted from each example either manually (e.g., by determining dough thickness, cheese type, serving temperature, toppings, etc.) or using some automatic feature extraction mechanism. The types of features, machine learning model (Support Vector Machine, Random Forest, Deep Convolutional Neural Network, etc.), hyper-parameters (regularization constant or number of convolutional blocks in deep convolutional network, etc.), evaluation metrics (area under the ROC curve or accuracy, etc.) and performance assessment protocol (k-fold cross-validation, random test split, etc.) are all decided by the machine learning engineer in the hopes of building a useful machine learning model (see Figure 1). Ideally, such a machine learning model, when deployed, should generalize well in the real world (i.e., be useful enough to tell us that a burnt pizza is not nice!). If a machine learning model cannot do this, then in line with the words of Richard P. Feynman, "It's wrong!"[1]. However, in academia, practical deployment of machine learning models, although desirable, is typically preceded by submission of a research article with the claim that its performance is better or at least as good as any previously published methods. Inappropriate or incomplete analyses of machine learning techniques, especially their performance assessment, can have serious consequences [2], [3]. It leads to difficulty in reproducibility, can stunt the growth of the field, or prove to be incredibly frustrating for those trying to improve existing results [4], [5]. In this article, we present ten not-to-do things that can introduce bias in performance evaluation leading to over-optimistic results. Our objective is to alert researchers to some caveats in developing, publishing, reviewing and, and most importantly, deploying machine learning models[2]. We first give a point, discuss its implications and then give lessons that should be considered in developing machine learning models.

---

[1] Richard P. Feynman on the Scientific Method: "It doesn't matter how beautiful your theory is, it doesn't matter how smart you are. If it doesn't agree with experiment, it's wrong!".

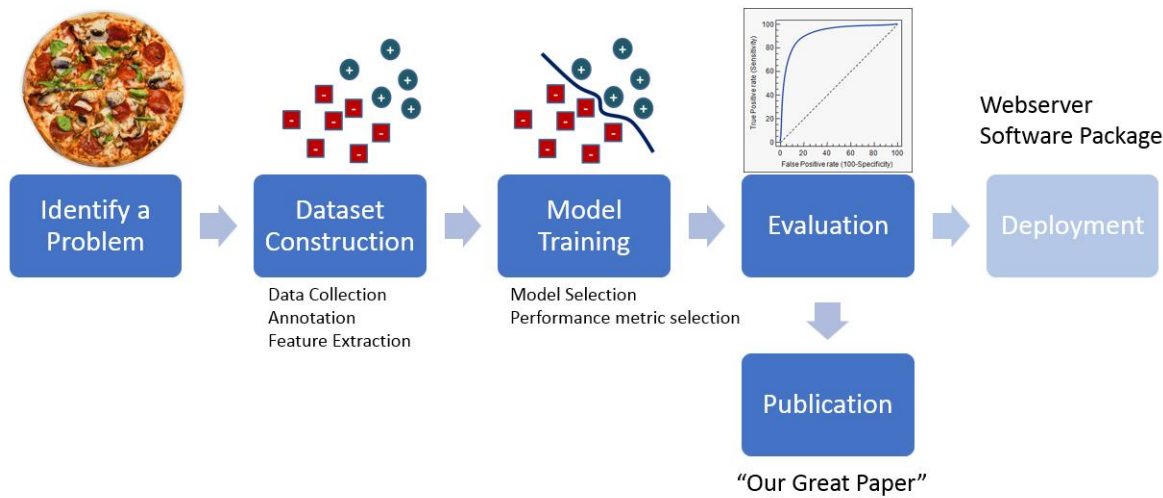

*Figure 1: Typical flow of machine learning development process*

1. **Use an uninformative or irrelevant accuracy metric**

    In order to assess the performance of a machine learning model, different accuracy metrics can be used. These include, accuracy, balanced accuracy, area under the receiver operating characteristic or precision-recall curve, root mean squared error, etc. [7], [8]. It is important to note that different metrics are appropriate under different situations [9]. For instance, percentage classification accuracy is appropriate if classifier decisions are generated at the optimal threshold level, class sizes are balanced, and false positives are as bad as false negatives. To illustrate this point further, let's consider a classification problem of predicting whether an individual has a disease or not using machine learning. If the "background" probability of having that disease is only 2%, then a practically useless classifier that classifies everything as negative will get an accuracy of 98%! Area under the receiver operating characteristic curve should not be used in imbalanced datasets or when the actual number and not the percentage of false positives is important [10], [11]. Another example would be the use of root mean squared errors in regression whose absolute value is meaningless unless the magnitude and range of the target values is known. It is very easy to fall victim to choosing an inappropriate performance assessment metric. Using an inappropriate measure that is not aligned with the intended use will lead to expectations that cannot be met by a classifier in practical deployment.

    Using a performance metric that is irrelevant to the problem domain may have a similar impact. Nowadays, one comes across many machine learning based algorithms for problems from a variety of domains. However, several machine learning researchers, while presenting their method for a problem from another domain, present performance analysis in a way that is not understandable for the domain expert. For example, a biologist might be interested in a system that would help his/her search for a protein with a certain trait to be narrowed down from millions to a handful. When a machine learning

---

[2] The idea for the title of the paper is taken from an analogous paper in the parallel computing domain [6]. We were also inspired by the "Ten Simple Rules" series of PLoS Computational Biology.

expert develops such a system, the analysis should consist of performance metrics that are useful and interpretable for the domain expert. In the previously mentioned example, a ranking analysis of the system's predictions might be more useful for a biologist than claims about accuracy or precision of the system. An analysis comprising of just the textbook measures like accuracy, precision, recall, sensitivity etc. might be useful from a machine learning expert's perspective but will not help in any practical use of the system.

**Lesson(s):**
a. Identify an appropriate performance metric based on the machine learning problem
b. Justify the use of a performance metric for a machine learning project
c. Choose problem-specific performance metrics that can help a user decide the usefulness of your method in their problem.

2. **Inappropriate model selection strategy**

Ronald Coase is attributed to have said *"If you torture the data long enough, it would confess to anything!"*. Machine learning models with high capacity (e.g., Support Vector Machines with the RBF kernel or deep neural networks) are prone to overfitting if not trained carefully [12], [13]. Given enough capacity, (for example, enough neurons and deep structure in case of neural networks), a method can memorize any data distribution. Also, if not done appropriately, model selection may introduce overfitting [14]. Feature selection performed on the basis of a the entire dataset can also give rise to significant bias in a machine learning model [15]. In the machine learning development lifecycle, direct or indirect use of test labels is not allowed. However, forgetting this requirement and optimizing hyperparameters using the test data for validation instead of a dedicated validation set from the training data can help produce overly optimistic results (see Fig 2 for this erroneous lifecycle). Using the same data for performance evaluation as that used for hyperparameter optimization may result in overestimation of the true ability of a system to generalize.

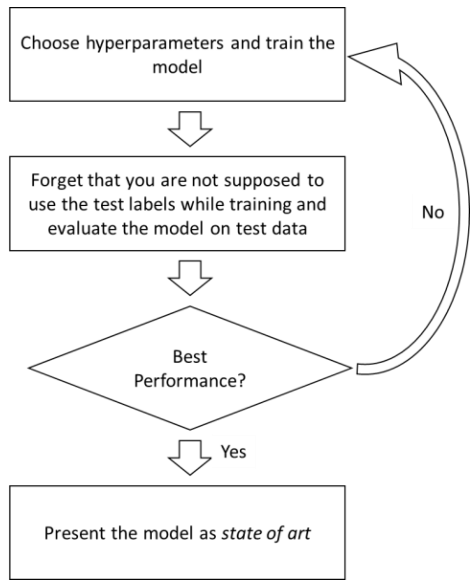

*Figure 2: Machine Learning lifecycle optimized for giving "state of the art" results*

Typically, researchers in machine learning present cross-validation results to demonstrate the performance of their methods. For example, one may choose a real-world dataset and decide to present K- fold cross-validation results. In K- fold cross validation, a dataset is divided into K parts, out of which, K-

1 are used for training and the remaining one for testing/ validation [16]. This process is repeated for all K partitions. Now consider the following cross-validation scheme:

```
Initialize my_model, results=[]
for fold in Folds:
    my_model.train(fold.train_data)
    p=my_model.evaluate_performance(fold.test_data)
    results.append(p)
final_score=average(results)
```

That is, instantiate a single *model* instance and then use the same instance for all folds, forgetting to re-initialize it in each fold. Hence, getting a train/ test overlap for folds subsequent to the first fold leading to overvaluation of performance results. Since the model is already trained over examples that will be a part of test/validation set in all the cross-validation iterations after the first one, the performance evaluation no longer remains meaningful as the model is very likely to have memorized the examples already and is likely to make accurate predictions.

**Lesson(s):**
  a. Use a validation set that is part of the training data for hyperparameter optimization.
  b. Do not make any direct/ indirect use of test labels for fair performance evaluation.
  c. Be very careful while writing code for performance evaluation. Small logical bugs can lead to big performance estimation errors.

3. **Ignore the fact that examples may not be independent of each other**

As mentioned earlier, it is very common in machine learning studies to use subsets of the same dataset for training and testing/validation for performance evaluation. A random splitting strategy is typically used for the purpose. However, it is possible that examples in a dataset are not independent of each other, i.e., there may be closely related examples in a dataset. In such cases, random splitting may yield train/test pairs with significant overlap due to presence of closely related examples in the two sets (see Figure 3).

As an example, consider a dataset comprising of protein data. If there exists close homology among examples, i.e., they share considerable sequence similarity and if random splitting is used, some of the homologs will be used for training and others for testing/validation, hence leading to overestimation of performance [17]. In such cases, cross-validation with a *Leave One Group Out* protocol should be used in which examples in the same group are not split across train and test sets.

**Lesson(s):**
  a. Study the intended use case of the machine learning problem and its domain to identify whether examples in a given data set are independent and identically distributed or not.
  b. Make sure there is no significant relatedness between train and test/validation sets to avoid overly optimistic estimates of performance.
  c. Datasets should be analyzed in depth to avoid overlaps between train and test sets. Domain experts should be consulted for the purpose.

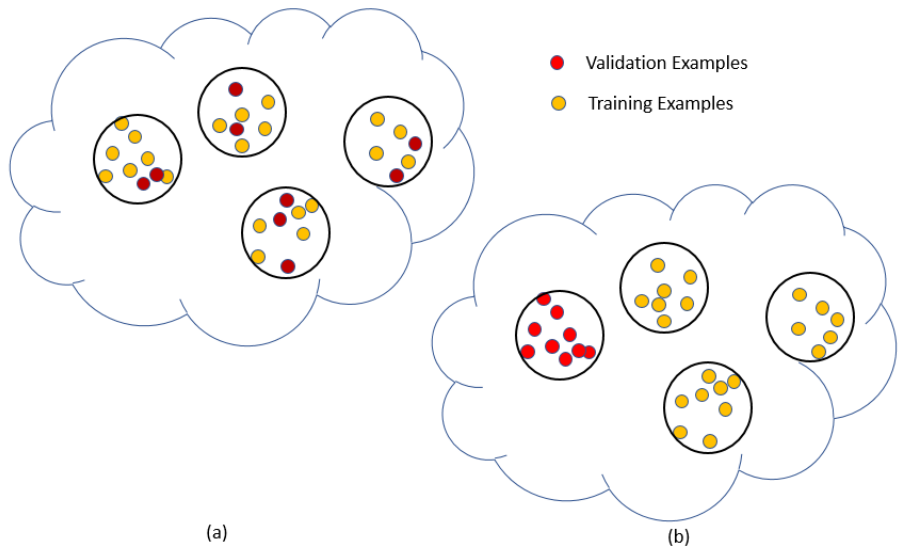

*Figure 3: Examples in machine learning task may not be independent of each other (indicated by circles). For examples, some examples in a medical diagnosis task may come from one subject whereas the machine learning task may require generating predictions for new subjects. If validation is done by random splitting as (a), then an overly optimistic performance assessment as opposed to the more realistic, leave one group out cross validation (b).*

4. **Do not compare with simple baseline classifier**

   Instead of straightaway opting for a very complex model for a problem, simpler methods should be tried first to have a baseline for performance evaluation. A method cannot be evaluated fairly if there is no well-established baseline for comparison. For example, consider the following scenario: A simple linear classifier gives 95% accuracy on a certain problem and a very complex and deep model manages to give a mere improvement of 1-2%. Also, complex models are more prone to overfitting and are resource-hungry. Presenting 97% accuracy using a very complex model without mentioning the fact that the simpler method was not too far behind in performance is another way of overselling a method. The principle of Occam's razor when applied on Machine Learning, demands that if the performance is roughly the same, the simpler of the two methods should be picked [18]. Also, when deep, multi-stage or ensemble prediction systems are developed, the use of depth, multiple stages and classifiers should be justified in light of the problem [19]. Explanation of what different components add to the system and how the system's performance would deteriorate if they are not used, should be presented. Justification of choosing complex models over simpler ones should also be given.

   **Lesson(s):**
   a. Before trying a complex solution, establish a baseline using simple methods.
   b. In case of a tie in performance, simpler is better.
   c. Justify the use of all techniques and steps in the model.

5. **Compare your model with un-optimized versions of other models or ones that have been trained using different data**

   A method may look exceptional in comparison to others if its optimized version is compared with un-optimized versions of others. For example, while performing comparative analysis, one chooses the best hyperparameter configuration for one's model using techniques like grid search, whereas, default are used for other methods. A comparison would be fair only if the same level of optimization is used for all the methods. Also, results for two models are not comparable if they are not trained and tested over the same data in the same experimental conditions. If cross-validation schemes, datasets, and other experimental conditions are not kept same across the methods, the comparison cannot be deemed fair. An example is the study of Weinberg et al. [20], that demonstrates the importance of removing experimental biases favoring some methods before performing comparison among different methods. If such biases are not removed, we may get exaggerated performance estimates for some methods.

**Lesson(s):**
   a. Comparison among methods should be performed under the same experimental conditions.
   b. Numbers from comparative analyses are meaningless unless all the experimental biases favoring certain methods are addressed.

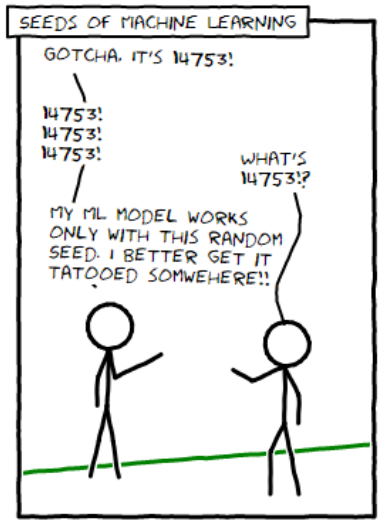

6. **Present your paper in a way that doesn't allow reproducibility**

   Many fields of science are facing a reproducibility crisis [21], and machine Learning is one of the affected areas [4]. In a recent survey presented at ICML by Joelle Pineau, the majority of researchers in machine learning agree that there is a significant reproducibility crisis in this field [22]. There is an agreement in machine learning community on the importance of open availability of machine learning software [23]. However, many published studies do not provide code/software. Re-implementation of a system from scratch is a tedious task, and the absence of complete implementation details including hyperparameter settings, preprocessing, etc., make the task even more difficult. And hence, even if the method performs well in practice, it cannot be employed by others, and evaluating its performance on other datasets becomes challenging or even impossible. Sandve et al. have recently given a list of points on how to ensure reproducibility in computational research [24]. Similarly, Henderson et al. have outlined the need and measures for reproducible research in reinforcement learning [25].

   **Lesson(s):**
   a. Researchers should make code used in studies publicly available.
   b. If the code is not made public, sufficient implement details should be given so that the system can be implemented by other researchers.

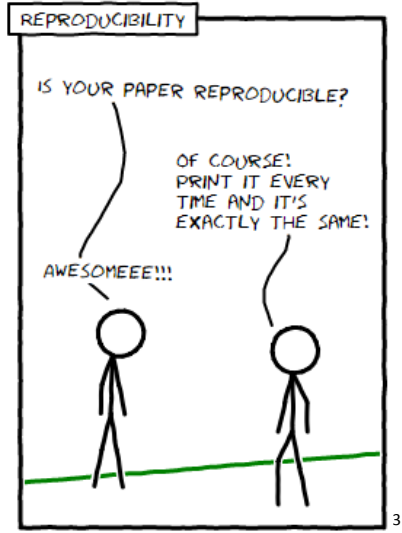

7. **Do NOT analyze the sensitivity of your model to changes in data, hyper-parameter values or randomness**

   A good machine learning model is robust and does not collapse with small changes in the experimental settings such as hyperparameters. Although a model that is very sensitive to small changes in the setup may perform well on a given dataset, it is not expected to generalize well and is very likely to fail in real world applications. Models that are drastically affected due to sources of randomness like weight initialization, validation set selection, etc. or small perturbations in the experimental conditions also suffer from poor generalization. A real world example of such scenarios is the one pixel attack on deep neural networks [26], where a single pixel change in the input image alters the output of the system and results in a wrong class prediction. A similar study of fooling deep models with small perturbations has been presented in [27].

   **Lesson(s):**
   a. Perform rigorous sensitivity analysis before deploying the system.
   b. Test your system for robustness to noise before deploying it for real world applications.

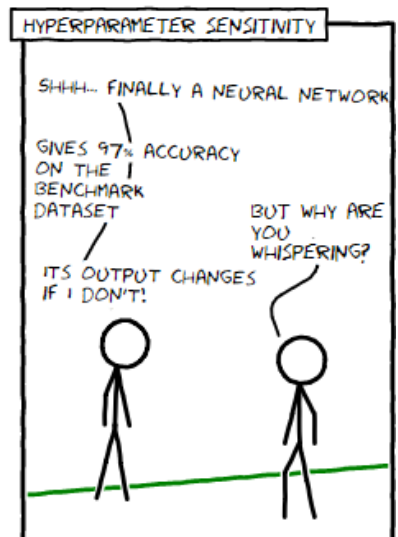

8. **Use statistical tests even when their underlying assumptions are not met**

---

[3] The XKCD style comics used in the paper have been created by the authors using http://cmx.io.

*"There are three kinds of lies: lies, damned lies, and statistics." (Mark Twain)*

Every statistical test has some underlying assumptions about the distribution of the data on which they are applied, for example, IID (Independent and Identically Distributed), normality, linearity, equality of variance etc [28]–[30]. Not many machine learning datasets fulfill these requirements. Statistical tests should not be used without verifying their applicability. Using such tests without ensuring their eligibility just because they produce big numbers may also lead to raised expectations from the method.

**Lesson(s):**
   a. Before using statistical tests, make sure the underlying assumptions are fulfilled.
   b. Justify the use of each test employed for evaluation.

9. **Do NOT worry about what the is model learning**

   Since the advent of practical deep learning and automatic feature extraction, many machine learning tasks have been simplified. However, in several cases, unusual properties of data may lead to extraction of irrelevant features and hence decisions being made on wrong criteria. For example, consider a scenario where the task is to use machine learning for identifying cats and dogs in images. Suppose all the cat images were captured in daylight and the dog images were taken at night. During training the deep model might start learning overall brightness and light as a discriminative feature. Validation on the same set will produce high performance scores. But in real applications, it would fail on cat images captured at night and dog images taken in bright light settings. Many such incidents have been reported in the literature over the years; some as cautionary tales, others as actual events [31].

   In deep learning, especially convolutional networks, it is important to analyze the weights the net is learning. Visualizing feature maps in convolutional nets can give an insight on what kind of features is the network using for making decisions [32]. If the feature maps do not look like they are extracting any useful information, the model should be tested in varying experimental settings and data to ensure proper learning. Several other efforts to make AI explainable [33] have been proposed in the literature including LIME, SHAP and others [34]–[36].

   **Lesson(s):**
      a. Learned weights of a trained models should be analyzed to ensure that the model is learning useful information.
      b. Feature map visualization for CNNs should be performed to confirm proper learning of the models.

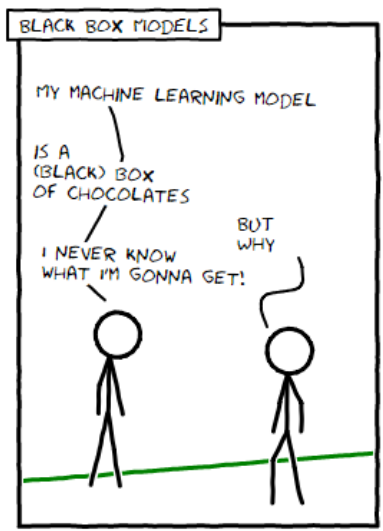

10. **Use buzzwords and pretty plots to whip your readers into submission**

    *"If you can't convince them, confuse them". (Harry S. Truman)*

A study should be present in as simple words as possible. Unnecessary use of technical jargon is both intimidating and distracting. Too many difficult words confuse the readers and keep them busy in deciphering the technical terminologies while keeping their attention away from the essence of the study. Very complex plots, and big and scary words may confuse the readers. In addition to that, focusing only on the trendy machine learning models without proper baseline comparisons can also fall into this [37].

**Lesson(s):**
   a. Use simple and easy-to-understand language while presenting your work.
   b. Present your study in a way that minimal preliminary study is required to understand your work. All that is needed to understand your research article should be present in there.

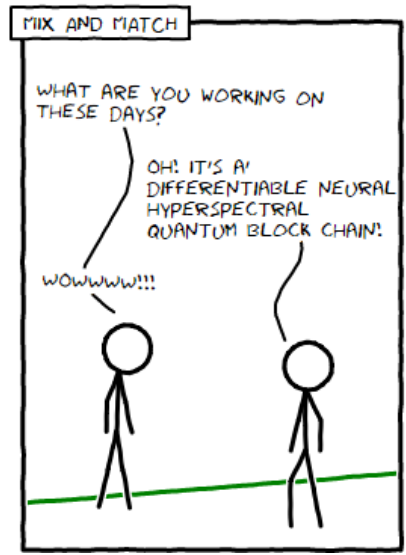

## Conclusions

In this paper, we presented some of the factors that cause a machine learning model to look more promising than it actually is. Vigilance of reviewers, and stringent publication criteria of journals and conferences can help avoid these problems. Awareness needs to be raised among researchers in the field to address these issues.

## References

[1]   "Hype Cycle for Data Science and Machine Learning, 2018." [Online]. Available: https://www.gartner.com/doc/3883664/hype-cycle-data-science-machine. [Accessed: 11-Dec-2018].
[2]   A.-L. Boulesteix, "Ten Simple Rules for Reducing Overoptimistic Reporting in Methodological Computational Research," *PLOS Comput. Biol.*, vol. 11, no. 4, p. e1004191, Apr. 2015.
[3]   Z. C. Lipton and J. Steinhardt, "Troubling Trends in Machine Learning Scholarship," *ArXiv180703341 Cs Stat*, Jul. 2018.
[4]   M. Hutson, *Artificial intelligence faces reproducibility crisis*. American Association for the Advancement of Science, 2018.
[5]   R. J. LeVeque, I. M. Mitchell, and V. Stodden, "Reproducible research for scientific computing: Tools and strategies for changing the culture," *Comput. Sci. Eng.*, vol. 14, no. 4, pp. 13–17, Jul. 2012.
[6]   D. H. Bailey, *12 WAYS TO FOOL THE MASSES WHEN GIVING PERFORMANCE RESULTS ON PARALLEL COMPUTERS*. ASFRA VOORHAVEN 33, 1135 BL EDAM, NETHERLANDS, 1991.
[7]   M. Sokolova and G. Lapalme, "A systematic analysis of performance measures for classification tasks," *Inf. Process. Manag.*, vol. 45, no. 4, pp. 427–437, Jul. 2009.

[8] E. Alpaydin, *Introduction to Machine Learning*, Third edition edition. Cambridge, Massachusetts: The MIT Press, 2014.
[9] N. Japkowicz and M. Shah, *Evaluating Learning Algorithms: A Classification Perspective*. Cambridge University Press, 2011.
[10] T. Saito and M. Rehmsmeier, "The Precision-Recall Plot Is More Informative than the ROC Plot When Evaluating Binary Classifiers on Imbalanced Datasets," *PLoS ONE*, vol. 10, no. 3, Mar. 2015.
[11] J. Brabec and L. Machlica, "Bad practices in evaluation methodology relevant to class-imbalanced problems," *ArXiv181201388 Cs Stat*, Dec. 2018.
[12] N. Srivastava, G. Hinton, A. Krizhevsky, I. Sutskever, and R. Salakhutdinov, "Dropout: a simple way to prevent neural networks from overfitting," *J. Mach. Learn. Res.*, vol. 15, no. 1, pp. 1929–1958, 2014.
[13] A. Ben-Hur, C. S. Ong, S. Sonnenburg, B. Schölkopf, and G. Rätsch, "Support Vector Machines and Kernels for Computational Biology," *PLoS Comput Biol*, vol. 4, no. 10, p. e1000173, Oct. 2008.
[14] G. C. Cawley and N. L. C. Talbot, "On Over-fitting in Model Selection and Subsequent Selection Bias in Performance Evaluation," *J. Mach. Learn. Res.*, vol. 11, no. Jul, pp. 2079–2107, 2010.
[15] C. Ambroise and G. J. McLachlan, "Selection bias in gene extraction on the basis of microarray gene-expression data," *Proc. Natl. Acad. Sci.*, vol. 99, no. 10, pp. 6562–6566, May 2002.
[16] S. Arlot, A. Celisse, and others, "A survey of cross-validation procedures for model selection," *Stat. Surv.*, vol. 4, pp. 40–79, 2010.
[17] W. A. Abbasi and F. U. A. A. Minhas, "Issues in performance evaluation for host–pathogen protein interaction prediction," *J. Bioinform. Comput. Biol.*, vol. 14, no. 3, p. 1650011, Jan. 2016.
[18] P. Domingos, "The role of Occam's razor in knowledge discovery," *Data Min. Knowl. Discov.*, vol. 3, no. 4, pp. 409–425, 1999.
[19] C. Kaynak and E. Alpaydin, "Multistage cascading of multiple classifiers: One man's noise is another man's data," in *ICML*, 2000, pp. 455–462.
[20] M. Wainberg, B. Alipanahi, and B. J. Frey, "Are Random Forests Truly the Best Classifiers?," *J. Mach. Learn. Res.*, vol. 17, no. 110, pp. 1–5, 2016.
[21] M. Baker, "Is there a reproducibility crisis? A Nature survey lifts the lid on how researchers view the 'crisis rocking science and what they think will help," *Nature*, 26-May-2016. [Online]. Available: http://link.galegroup.com/apps/doc/A454730385/AONE?sid=googlescholar. [Accessed: 16-Dec-2018].
[22] J. Pineau, "The ICLR 2018 Reproducibility Challenge," presented at the ICML workshop on Reproducibility in Machine Learning, 2018.
[23] S. Sonnenburg *et al.*, "The Need for Open Source Software in Machine Learning," *J. Mach. Learn. Res.*, vol. 8, no. Oct, pp. 2443–2466, 2007.
[24] G. K. Sandve, A. Nekrutenko, J. Taylor, and E. Hovig, "Ten Simple Rules for Reproducible Computational Research," *PLOS Comput. Biol.*, vol. 9, no. 10, p. e1003285, Oct. 2013.
[25] P. Henderson, R. Islam, P. Bachman, J. Pineau, D. Precup, and D. Meger, "Deep Reinforcement Learning that Matters," *ArXiv170906560 Cs Stat*, Sep. 2017.
[26] J. Su, D. V. Vargas, and S. Kouichi, "One pixel attack for fooling deep neural networks," Oct. 2017.
[27] S.-M. Moosavi-Dezfooli, A. Fawzi, and P. Frossard, "DeepFool: a simple and accurate method to fool deep neural networks," *ArXiv151104599 Cs*, Nov. 2015.
[28] J. Demšar, "On the appropriateness of statistical tests in machine learning," in *Workshop on Evaluation Methods for Machine Learning in conjunction with ICML*, 2008, p. 65.
[29] S. L. Salzberg, "On Comparing Classifiers: Pitfalls to Avoid and a Recommended Approach," *Data Min. Knowl. Discov.*, vol. 1, no. 3, pp. 317–328, Sep. 1997.
[30] J. Demšar, "Statistical Comparisons of Classifiers over Multiple Data Sets," *J Mach Learn Res*, vol. 7, pp. 1–30, Dec. 2006.
[31] gwern, "The Neural Net Tank Urban Legend - Gwern.net," 20-Sep-2011. [Online]. Available: https://www.gwern.net/Tanks. [Accessed: 16-Dec-2018].
[32] M. D. Zeiler and R. Fergus, "Visualizing and Understanding Convolutional Networks," in *Computer Vision – ECCV 2014*, 2014, pp. 818–833.
[33] D. Gunning, "Explainable artificial intelligence (xai)," *Def. Adv. Res. Proj. Agency DARPA Nd Web*, 2017.
[34] J. Chen, L. Song, M. J. Wainwright, and M. I. Jordan, "Learning to Explain: An Information-Theoretic Perspective on Model Interpretation," *ArXiv Prepr. ArXiv180207814*, 2018.

[35] M. T. Ribeiro, S. Singh, and C. Guestrin, "'Why Should I Trust You?': Explaining the Predictions of Any Classifier," *ArXiv160204938 Cs Stat*, Feb. 2016.
[36] S. Lundberg, *A unified approach to explain the output of any machine learning model.: slundberg/shap*. 2019.
[37] Hypergiant, "Is Neural Network Hype Killing Machine Learning?," *Hypergiant*, 07-Sep-2018. .